# Future Artificial Intelligence tools and perspectives in medicine

Ahmad Chaddad[a,*], Yousef Katib[b], Lama Hassan[a]

[a]School of Artificial Intelligence, Guilin University of Electronic Technology, Guilin 541004, China

[b]Department of Radiology, Taibah University, Al-Madinah 42353, Saudi Arabia

[*]Correspondence to Ahmad Chaddad, PhD,

School of artificial Intelligence, Guilin University of Electronic technology,

Guilin, Guangxi 541004, China

Tel.: +1-514-619-0751 or +86-150-7730-5314;

Email: ahmadchaddad@guet.edu.cn or ahmad.chaddad@affiliate.mcgill.ca

**Purpose of review**

Artificial intelligence (AI) has become popular in medical applications, specifically as a clinical support tool for computer-aided diagnosis. These tools are typically employed on medical data (i.e., image, molecular data, clinical variables, etc.) and used the statistical and machine learning methods to measure the model performance. In this review, we summarized and discussed the most recent radiomic pipeline used for clinical analysis.

**Recent findings**

Currently, limited management of cancers benefits from artificial intelligence, mostly related to a computer-aided diagnosis that avoids a biopsy analysis that presents additional risks and costs. Most AI tools are based on imaging features, known as radiomic analysis that can be refined into predictive models in non-invasively acquired imaging data. This review explores the progress of AI-based radiomic tools for clinical applications with a brief description of necessary technical steps. Explaining new radiomic approaches based on deep learning techniques will explain how the new radiomic models (deep radiomic analysis) can benefit from deep convolutional neural networks and be applied on limited data sets.

**Summary**

To consider the radiomic algorithms, further investigations are recommended to involve deep learning in radiomic models with additional validation steps on various cancer types.

**Introduction**

Artificial intelligence (AI) is a popular term used in many fields for different goals. Currently, AI has potentially boosted health care in terms of clinical practices (1,2). These AI tools are mainly being used for many clinical tasks like diagnosis, treatment, and prognosis (3,4). For example, AI with radiomic models offers a non-invasive method to assess clinical predictions. Given cancer as an example, a standard radiomics is a quantitative technique that extracts features from medical data (i.e., image) and uses these features as input to conventional classifier models (5–10). However, the radiomic analysis can provide a wide range of features from labeled regions/areas that affect the classifier model and lead to model overfitting with a bias in results (11,12). This strategy is well performed on many cancer entities (13–16). Understanding informative features to be used as the input features to classifier models are still limited (17).

Many studies proposing computer-aided diagnostic (CAD) systems to detect and identify tumours on medical images using either radiomics and machine learning or deep learning methods have grown considerably due to increasing interest in AI techniques and their medical applications (1), like the detection (18,19), segmentation (20), classification (21–26) and survival analysis (27). Specifically, convolutional neural networks (CNNs) offer a dynamic model in which the predictive features are extracted directly from images (e.g., MRI, CT-scan, PET, etc.) in an iterative process and avoid the handcraft features. More precisely, the deep CNNs consist of convolutional and pooling layers representing feature extraction and feature selection, respectively. During the CNN training, these features changeable to find the highest informative features in a fully connected layer to then use as input to the classifier (i.e., Softmax) (28,29).

Moreover, CNN has revolutionized image analysis since its remarkable win in the image recognition contest ILSVRC (ImageNet Large Scale Visual Recognition Challenge) in 2012 (30) with many physicians and researchers who have attempted to harness the power of AI (or, more precisely, CNN) for clinical applications. This happened because it avoids the need to generate detailed features by craftsmanship. One example of strong influence is the efficacious classification of dermoscopy images (31,32).

Nowadays, AI became as a key driver of the conversion of health care to precision medicine. For example, it has achieved a level of precision in interpreting mammograms for breast cancer screening (i.e., area under the ROC curve (AUC) for AI of 0.840, compared with 0.814 for physicians) (33). In 2018, the food and drug administration (FDA) approved the cloud-based Arteries imaging platform as a tool to help physicians track tumors base on computed tomography scans and MRI of lung and liver cancer patients. Also, it was designated as a breakthrough device by the FDA that uses deep learning-based AI algorithms to detect, diagnose and predict certain cancer types[1]. Many examples have been demonstrated the usefulness of deep CNNs. For instance, CNNs have been able to detect colonic polyps or enlarged lymph nodes in computed tomography images (34). Likewise, a deep learning system has been used to detect cancer areas in whole slide images of radical prostatectomy specimens and automatically assign the Gleason score with an accuracy of 0.70 (which is superior to general pathologists) (35,36). Unfortunately, this process looks like a black-box and rarely offered an interpretation related to hidden/shallow layers (27,37). Despite the massive works involving the CNN models with radiomic analysis, overfitting results are still considered one of the deep learning and radiomic model challenges (38).

A combination approach of conventional radiomic and deep CNN is recently proposed to reduce the bias and overfitting (27,39). It approached the understanding ability of information flow through the hidden layers in CNN model. This is likely the term of *explainability,* which aims to show the most decision model's informative features (40). This review paper describes the studies related to the AI with radiomic with the main steps of standard radiomic pipeline and deep radiomic models in clinical applications. Therefore, we will identify the most common radiomic model and determine the clinical practices' advanced steps. To simplify the technical terms, we report the keywords in **Table 1**.

---

[1] https://www.fda.gov/medical-devices/software-medical-device-samd/artificial-intelligence-and-machine-learning-software-medical-device

**Table 1: Summary of keywords description**

| Keywords | Definitions |
|---|---|
| Deep learning | It can be considered a subset/type of machine learning that uses deep neural networks to train a computer to perform human tasks. |
| CNN | It is a type of deep learning technique known as deep convolutional neural networks (CNN). It is used for classification, automatic segmentation, detection, and feature generator. Usually, CNN is considered deep CNN since it has more than one layer. CNN architecture is a stack of convolutional and pooling layers, followed by a fully connected layer and classifier (i.e., Softmax). |
| Overfitting | If the predictive model does much better on the training set than on the validation set, we name it overfitting. |
| Convolutional layer | It uses filters that perform convolution operations. It is scanning the input image concerning its dimensions. The resulting output is called a feature map or activation map. |
| Pre-trained CNN network | Is a model trained by other dataset (created by someone else) to solve a problem (classification) that is similar to ours (normal versus cancer). |
| Entropy functions | It measures the disorder of the image texture. |
| Quantifiers functions | Use to measure the image feature (or texture) quantitatively. |
| Omics | Omics represent comprehensive approaches for molecular analysis from a biological sample. |

**AI-based radiomics**

The basic of radiomic pipeline consists of the following steps: Image acquisition based on scanner machine (MRI, CT-Scan, PET, etc.), segmentation of abnormal tissues known as regions of interest (ROI) or gross tumor volume (GTV), and computation of features to build a predictive model (41–45). With deep learning, the radiomic model could be represented by a CNN model that does not need to make the labeling and feature selections. With the advantage of deep CNN, the current radiomic model is based on a hybrid CNN model and standard radiomic. Figure 1 illustrates the development procedure of radiomic pipelines with a large number of medical data. We grouped the radiomic pipelines in three categories as follows:

*1. Standard radiomic pipeline*

It is the most popular radiomic model. It requires a semi- or fully automatic labeling/segmentation for ROI/GTV, computation of many image features (i.e., shape, histogram, texture; deep CNN features, etc.(46)) are derived from ROI/GTV images (47,48). Furthermore, essential features are selected based on their predictive impact (e.g., the importance of features, dominant features) or/and significance value (e.g., significant features). These features could be analyzed by significance test and/or Spearman correlation or combined to build predictive models. Such models were applied to many cancer types (13,15,43,46,49). This type of radiomic pipelines uses mainly the imaging features (e.g., shape, texture, deep features) with clinical variables (e.g., age, gender, treatment, survival, etc.).

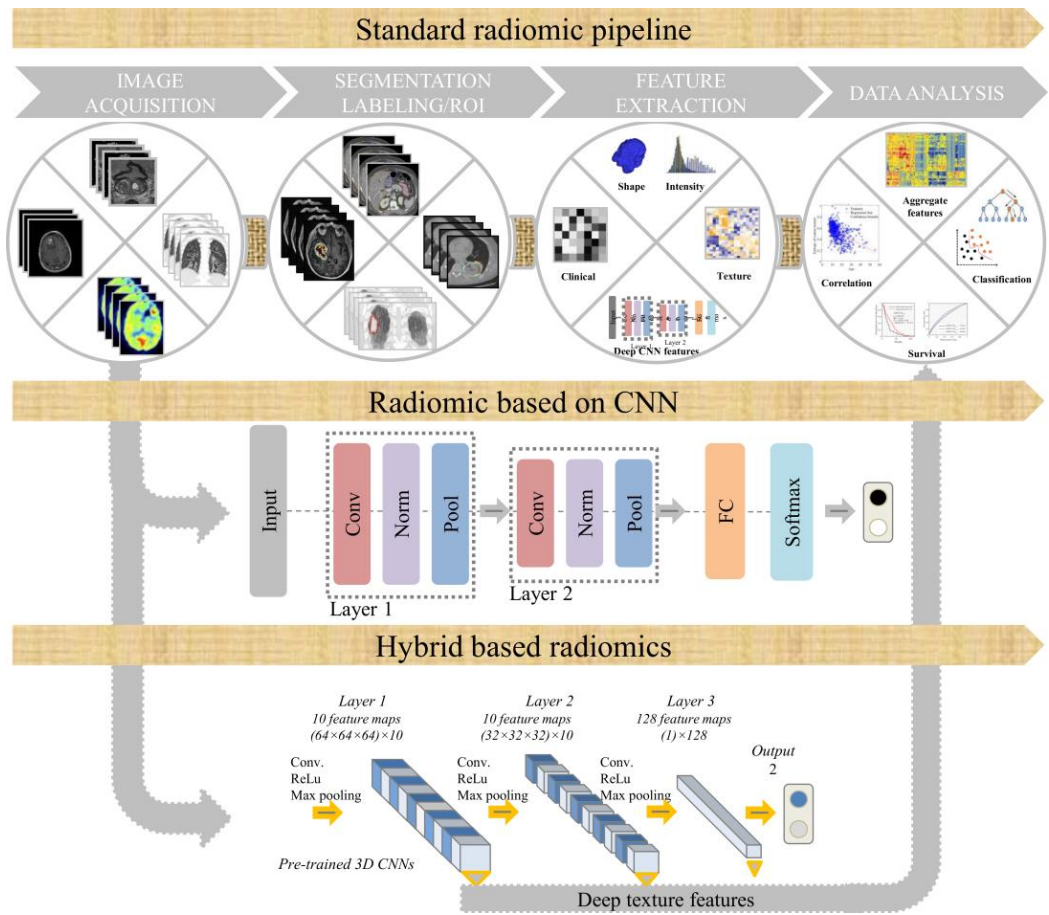

Figure 1: Three scenarios of AI with radiomic pipeline. 1) Standard radiomic: data acquisition using medical imagery equipment (e.g., MRI, CT-Scans, PET, Ultrasounds, etc.). Segmentation (labeling the regions of interest-ROI) using the manual (or semi-automatic) tools. Feature extraction from ROI images using the conventional shape and texture features in addition to the clinical variables. Data analysis using the statistical tests (e.g., Log-rank, Wilcoxon test) and predictive models (e.g., Logistic regression-LR, random forest-RF, support vector machine-SVM, etc.). 2) Radiomic based CNN models: Use the image directly without labeling of ROIs and predict the clinical tasks (survival, aggressiveness, etc.). 3) Hybrid based radiomics: consider the image and/or ROIs image as input to pretrained CNN model, and generate deep texture features from shallow/hidden layers to predict the clinical tasks using the machine learning techniques (e.g., LR, RF, SVM, etc.).

In this context, radiomic features could be used to predict the molecular variable and /or combined to build a predictive model based on radiogenomics and multi-omics data (16,50–58). In such radiomic pipelines, should manage some communications and collaborations. For example, the clinicians must ensure that the appropriate ROIs are being labeled for radiomic (or radiogenomics) analysis. Biological scientists may transfer the molecular features (i.e., genetic and proteomic markers) that will impact the clinical course of a patient. Biomedical scientists understand the scientific problem and decide what AI algorithm should be developed to make a consistent AI model for radiomics analysis. The discussion with statisticians would allow a quantitative approach to provide a theoretically and statistically significant solution.

2. *Radiomic based CNN*

In deep CNN, images are used as input to the CNN model to predict clinical tasks (e.g., survival, grade, treatment, etc.). Specifically, the CNN model looks like a "black-box" with complicated processing steps. For example, CNN consists of a frequent stack of convolution and pooling layers, followed by one or more fully-connected layers (59–61). Convolution layers generate spatial features from the image. These features represent the local image patterns and texture in initial and deep layers, respectively. For the radiomic model, two strategies could be considered, (a) deep CNN is used in transfer learning strategy (62–65). The basic idea of this strategy is to use a pre-trained CNN network and then use this network's features as the representation for learning a new task without re-training from scratch. Transferred features can be used directly as input to the new model or adapted to the new task via fine-tuning, (b) Use the CNN for feature extraction, whereas classifications could be performed with conventional classifier models (SVM, Random forest, etc.) due to the small training available data images. The most appropriate features in this scenario are related to the fully connected layer features (66).

3. *Hybrid based radiomics - deep radiomic analysis*

The hybrid radiomic model is based on deep CNNs and a conventional classifier for clinical tasks. For example, CNN is used to generate multiscale textures that encoded via quantifier functions. These function values are then used in a vector features as input to the conventional classifier. A recent example of a new radiomic model (names as deep radiomic analysis) uses popular deep CNNs to generate multi-scale texture encoded by entropy functions to predict the prostate cancer's Gleason score (37,67). These texture features are extracted from the hidden (shallow) layers of pretrained CNNs. This model is also known as radiomic based on deep texture features (39). The first kind of such radiomic model was proposed by Chaddad et al. in (68) to predict Alzheimer's disease. A similar pipeline model was also expanded to be adapted with 2D CNN models in generating deep texture features for predicting the GS of PCa patients (37). Such a radiomic model is proposed to avoid the problem of limited data to improve the model accuracy. This was proved also in recent work that uses the Gaussian Mixture Model (GMM) to encode the CNN features and generate a new radiomic signature to distinguish short-term from long-term survival of pancreatic ductal adenocarcinoma (27). Another example of radiomic model based on CNN and well-known quantifier functions was successfully applied to predict the survival outcome of recurrence brain tumor (69). One known problem is the lack of international collaboration to validate this type of radiomic models, then continuous collaboration and communication between institutions and involved the AI based radiomic approaches will be essential to groups studying deep radiomics analysis.

**Explainable AI (XAI)**

Explainability is a term used with AI systems related to experts (clinicians) in understanding the processing steps in the medical diagnosis system. More precisely, how the experts can involve in the AI systems and make the decisions. We limit the term of explainability in deep learning models that can be an open question for the future. Given CNN's hidden layers as an example of explainability, the neurons with their weight are considered as a black-box (40,70,71). A massive effort is currently trying to clarify the concept of XAI in the medical field (72–74). One of the efforts is represented by the recent works of Chaddad et al., which focus on understanding the flow of information in deep CNN during the training of data images for patients with a brain (69), prostate (37), and Alzheimer (39,68). Also, more details about deep learning with XAI methods are explained in (40,75).

Unfortunately, with these advanced algorithms, their clinical use in predicting individual risk is still limited (76). Unlike the current clinical settings, this perhaps is not a concern for manufacturing outlets, smart biomedical devices, a robot with surgical-aid workflow, etc. XAI with AI ethical standards lead to significant aspects in AI-based solutions.

**Future AI and key challenges**

AI with radiomic application is expected to follow the clinical demands closely and may require deep analysis to understand the underlying tissue characteristics and their relation with omics data (e.g., genomics, proteomics, DNA, RNA, etc.). Despite the advanced AI with radiomic models, unfortunately, a minority of the prior studies focus on AI-guided immunotherapy, which is recommended to be considered in further work. Also, the main challenge is how we validate the AI models for clinical practice with data derived from multi-center, large-sample, randomized-controlled clinical trials to achieve the goal of personalized medicine.

In this context, public datasets such as The Cancer Genome Atlas (TCGA) (77), The Cancer Imaging Archive (78), and The Quantitative Imaging Network (79) will boost the validation steps of the AI techniques to improve the radiomics stability. However, there is still the barrier of labeling the ROI that could be solved using the automatic mode based on deep learning (e.g., UNet, etc.) (80–83) and, recently, the application of domain adaptation (84). Moreover, deep learning models could perfectly work with extensive data analysis and improve the performance metrics. Here the clinicians (e.g., radiologists) must be involved in the AI development models suggested in XAI. We have to note the information security, privacy, and ethical issues that put the AI-radiomics analysis front many administrative steps that request more time. In the future, we see the radiomic models based on AI will consider the combination of many modalities (MRI, PET, CT, etc.) with big data (e.g., multi-omics) and clinical information to implement AI models for precision medicine.

**Conclusion**

In the review, we provide a brief overview of AI with radiomics tool in clinical applications. Specifically, this survey clarified the algorithm types used in radiomic analysis. However, AI's potential value with radiomics and explainability in clinical practice has not been fully investigated. For this reason, expanding the advanced AI tools and adjusting these techniques to be more clinically significant trends is recommended. With the progress of deep learning models applied to extensive public data, AI-based radiomics can facilitate clinical decision making non-invasively. Despite several AI scenarios with radiomic applications, the key challenge remains to ensure public access to comprehensive clinical and radiological related evidence leading to AI practice.

**Key points**

Artificial intelligence (AI) based radiomic is currently investigated to achieve the goal of personalized medicine.

Current studies showed that AI with radiomic analysis tools needs more validation steps to be involved in the clinical practice.

However, most of the available studies have a limited database, which is one of the most challenging for AI tools.

Further studies with a combination between standard radiomics and deep learning are needed to provide a generalizable AI tool.

**Acknowledgements**

None.

**Financial support and sponsorship**

Research supported by Foreign Young Talents Program (No. QN20200233001). The funding agency has no role in the conceptualization of the study, data collection and analysis, or the decision to publish these results.

**Conflicts of interest**

There are no conflicts of interest.